\documentclass{article}

\usepackage[preprint]{corl_2026}

\usepackage{multirow}
\usepackage{booktabs}
\usepackage{amssymb}
\usepackage{graphicx}
\usepackage{amsmath}
\usepackage{caption}
\usepackage{wrapfig}
\usepackage{enumitem}
\usepackage{array}

\usepackage{titlesec}
\titlespacing*{\section}{0pt}{0mm}{3pt}
\titlespacing*{\subsection}{0pt}{0mm}{3pt}

\usepackage{xcolor}
\definecolor{citecolor}{HTML}{0071BC}
\hypersetup{colorlinks,linkcolor={red},citecolor={citecolor}}

\title{DexCompose: Reusing Dexterous Policies for Multi-Task Manipulation with a Single Hand}
\author{
Dihong Huang\textsuperscript{*},
Zhenyu Wei\textsuperscript{},
Zhuxiu Xu\textsuperscript{*},
Yunchao Yao\textsuperscript{},
Sikai Li\textsuperscript{},
Mingyu Ding\textsuperscript{}
\\[0.5em]
\textsuperscript{} University of North Carolina at Chapel Hill
\\[0.5em]
\url{https://devon018.github.io/DexCompose-Webpage/}
}

\begin{document}
\maketitle

\renewcommand{\thefootnote}{\fnsymbol{footnote}}
\footnotetext[1]{This work was done during an internship at UNC.}

\begin{center}
    \vspace{-20pt}
        \includegraphics[width=\linewidth]{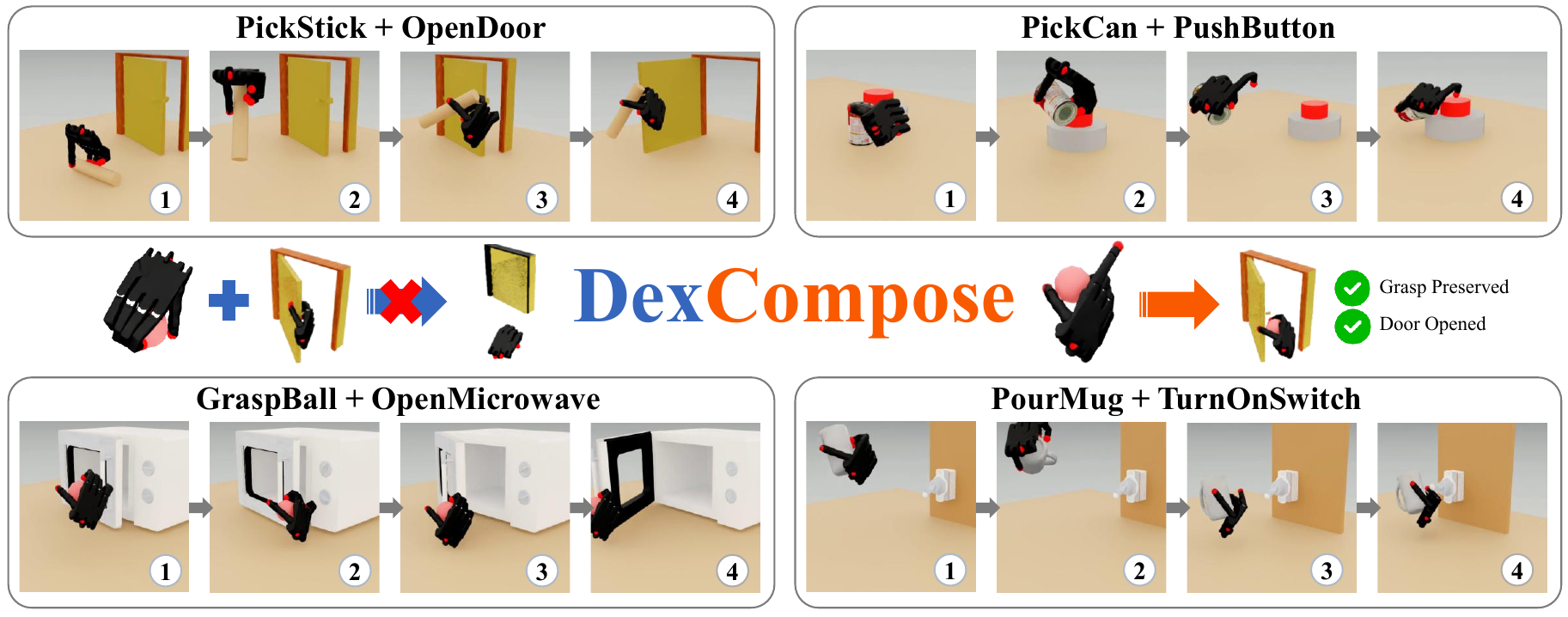}
    \vspace{-16pt}
    \captionof{figure}{DexCompose composes dexterous skills through role-aware finger ownership. By separating grasp preservation from downstream interaction with asymmetric residuals, the framework reduces destructive interference and enables robust multi-skill manipulation.}
    \vspace{2pt}
    \label{fig:teaser}
\end{center}

%===============================================================================
\begin{abstract}
Dexterous manipulation policies can solve individual skills, but composing them to perform multiple tasks with a single hand remains challenging.
Adding a new task on top of an existing manipulation skill often imposes conflicting demands on overlapping fingers and contact modes,
causing destructive interference between preserving an existing manipulation outcome and executing a new one.
We propose DexCompose, a role-aware residual composition framework that reuses pretrained dexterous policies for multi-task manipulation through explicit finger-level action ownership.
Given two pretrained full-hand policies, DexCompose first collects successful post-task states from the first skill and performs release tests over candidate finger masks to identify which fingers are necessary for maintaining the established skill state.
It then trains two asymmetric residual modules: a bounded residual stabilizer for task preservation, and a context-aware residual that adapts the frozen downstream policy only within the action subspace assigned to the new task.
We evaluate the framework on 16 composite dexterous manipulation tasks spanning four object-retention skills and four downstream interactions.
DexCompose achieves 77.4\% average composite success, 
demonstrating structural action ownership with dual residuals offers a promising direction for composing dexterous skills beyond conventional policy chaining.
% with ablations confirming that structural action ownership with dual residuals is more effective than conventional policy chaining or unmasked residual correction.
\end{abstract}

% Dexterous manipulation policies can solve individual skills, but composing them to perform multiple tasks with a single hand remains challenging. Adding a new task on top of an existing manipulation skill often imposes conflicting demands on overlapping fingers and contact modes, causing destructive interference between preserving an existing manipulation outcome and executing a new one. We propose DexCompose, a role-aware residual composition framework that reuses pretrained dexterous policies for multi-task manipulation through explicit finger-level action ownership. Given two pretrained full-hand policies, DexCompose first collects successful post-task states from the first skill and performs release tests over candidate finger masks to identify which fingers are necessary for maintaining the established skill state. It then trains two asymmetric residual modules: a bounded residual stabilizer for task preservation, and a context-aware residual that adapts the frozen downstream policy only within the action subspace assigned to the new task. We evaluate the framework on 16 composite dexterous manipulation tasks spanning four object-retention skills and four downstream interactions. DexCompose achieves 77.4\% average composite success, demonstrating structural action ownership with dual residuals offers a promising direction for composing dexterous skills beyond conventional policy chaining.


%===============================================================================
\section{Introduction}
\label{sec:introduction}

Dexterous manipulation often requires a hand to preserve the outcome of one skill while executing another. For example, after grasping a stick, a robot may need to use the same hand to push a door handle without dropping the object. Although grasping and pushing behaviors can each be learned by separate single-task policies, directly chaining these policies is unreliable: the downstream policy controls the full hand and may overwrite the finger motions required to maintain the grasp. This creates interference between two objectives that share the same action space: preserving the grasp while performing the new interaction task. A straightforward alternative is to train a separate policy for every composite task, but this scales poorly, requiring $n\times m$ policies for $n$ grasp-maintenance skills and $m$ downstream interactions. These limitations motivate a compositional approach that reuses pretrained single-task policies while explicitly managing how finger actions are allocated across objectives.

Recent work has shown the possibility of leveraging selective hand degrees-of-freedom (DoF) allocation to enable multiple dexterous interactions with a single hand.
MultiGrasp~\cite{li2024multigrasp} studies multi-object grasping by generating multi-object grasp proposals and learning a policy to execute them. Beyond grasping, DexMulti~\cite{jiang2026dexmulti} enables concurrent grasping and manipulation by decomposing interactions into reusable object-centric skills. HANDFUL~\cite{foong2026handful} learns resource-aware grasps by explicitly reserving fingers for downstream manipulation subtasks.
These methods suggest that complex dexterous interactions can be achieved by allocating hand resources across different manipulation objectives.
However, existing finger-aware approaches typically rely on predefined finger allocations or downstream task compositions during grasp generation, data construction, or policy training, making them difficult to scale across different task combinations.
Meanwhile, large-scale dexterous grasp datasets such as DexGraspNet, UniDexGrasp, and UniDexGrasp++~\cite{wang2023dexgraspnet,xu2023unidexgrasp,wan2023unidexgrasppp} have made reusable full-hand grasp controllers increasingly available. This motivates a question: \textit{can we compose pretrained dexterous manipulation policies by discovering and exploiting redundant hand DoFs across tasks?}

Under this perspective, we reframe task combinations as a resource-allocation problem at the embodiment level.
The key insight is that idle DoFs within a trained policy are a reusable resource for composing additional behaviors. Our approach should identify which action dimensions are necessary for preserving the first task's outcome, assign structural ownership of those dimensions, and restrict each subsequent policy to operate only within its allocated subspace. This reframing decomposes the problem into two sub-problems: \textbf{discovery}, identifying which action dimensions each policy actually needs, and \textbf{composition}, chaining different policies by ensuring each policy acts only through its allocated DoFs so that their objectives can be realized concurrently.

We propose DexCompose, a two-stage pipeline that addresses both sub-problems.
1) For discovery, we introduce finger attribution: we collect successful held-object states from the first task and identify finger-level redundancy through release tests. Specifically, we evaluate different finger masks by releasing subsets of fingers and observing whether the object remains stably retained. This process reveals which fingers are essential for maintaining the grasp. The task-specific mask is then selected to balance grasp stability with the dexterity needed for the downstream task. 2) For composition, we introduce dual residual stabilizer: one lightweight residual module preserves the held-object configuration against disturbances caused by base motion and redundant finger release, while the other provides task-aware compensation for the downstream policy under the motion constraints imposed by the maintained grasp.

We evaluate DexCompose on 16 composite manipulation tasks spanning four hold-and-retain tasks and four downstream interactions. Our method achieves $77.4\%$ average composite success, outperforming direct policy chaining by $74.3\%$ and the strongest baseline by $15.8\%$.
Results show that the dual residual stabilizer is the primary factor enabling successful composition, as it maintains stable grasp retention while adapting downstream actions under the constraints imposed by the maintained grasp. Finger allocation further improves composite success under the same training budget by reducing cross-task interference through structured action ownership.

In summary, our main contributions are as follows.
\begin{itemize}[leftmargin=12pt,itemsep=0pt,topsep=0pt,parsep=0pt]
    \item We introduce a post-hoc composition framework for pretrained dexterous manipulation policies, formulating composite manipulation as an action allocation problem where grasp preservation and downstream interaction must share the same high-DoF multi-fingered hand.

    \item We propose DexCompose, a role-aware residual pipeline that combines training-free finger attribution, explicit action ownership, and dual residual stabilization to compose two pretrained full-hand policies without policy training for the complete task pair.

    \item We evaluate DexCompose on 16 composite manipulation tasks and show that it substantially improves composite success over policy chaining and unmasked residual baselines, highlighting a scalable path toward reusable dexterous manipulation that can flexibly combine learned skills.
\end{itemize}

\section{Related Work}
\label{sec:related work}
% \vspace{-2mm}
\noindent \textbf{Dexterous policy composition.}
Prior work on dexterous policy composition mainly studies three directions: learning reusable dexterous controllers, sequencing them for long-horizon manipulation, and allocating fingers to enable multi-object interactions with a single dexterous hand.
A large body of work learns robust reusable dexterous policy via reinforcement learning, imitation learning, and generative or diffusion policies~\cite{popov2018data,rajeswaran2018learning,andrychowicz2020learning,akkaya2019solving,qin2022dexmv,mandikal2021dexvip,arunachalam2023dime,jiang2025dexmimicgen,chi2023diffusionpolicy,ze2024dp3,liang2025dexhanddiff}.
Multiple benchmarks and grasp-generation works scale dexterous manipulation across objects, articulated categories, cluttered scenes, and hand morphologies~\cite{bao2023dexart,wang2023dexgraspnet,xu2023unidexgrasp,zhang2025dexgraspnet2,wei2026one,zhang2026unidex,zhao2025dexh2r}.
The second line explicitly studies dexterous skill sequencing.
SequentialDexterity~\cite{chen2023sequentialdexterity} chains multiple dexterous policies by learning transition feasibility and selecting compatible policies across interaction stages.
Related long-horizon manipulation approaches also leverage skill priors, behavior primitives, or retrieval from prior data to reduce exploration and reuse previous experience~\cite{li2026coordex,pertsch2021accelerating,singh2021parrot,nasiriany2022maple,nasiriany2023sailor}.
The third direction exploits multi-finger hands for multi-object or multifunctional manipulation.
MultiGrasp~\cite{li2024multigrasp}, SeqMultiGrasp~\cite{he2025seqmultigrasp}, and SeqDiffuser~\cite{lu2025grasping} study multi-object grasp acquisition, while DexMulti~\cite{jiang2026dexmulti} and HANDFUL~\cite{foong2026handful} focus on multi-stage dexterous manipulation and future-aware grasping.
In contrast, our work approaches dexterous manipulation from the perspective of pretrained policy reuse, enabling the composition of two policies at inference time without modifying the original controllers.

\noindent \textbf{Residual policy learning.}
Residual policy learning improves an existing controller by learning an additive correction on top of a prior action.
Early residual formulations combine an imperfect policy with a learned residual, improving robustness under long horizons, partial observability, model mismatch, and sparse rewards~\cite{silver2018residual,ranjbar2021residual,shi2021proactive}.
Residual reinforcement learning further applies this idea to robot control by superimposing a learned residual action on a trained controller, where the controller governs the primary behavior while RL compensates for modeling errors and complex contact dynamics~\cite{johannink2019residual}.
Subsequent work improves upon priors by incorporating demonstrations, human commands, learned skills, or imitation policies, enabling residual learning for a wide range of robotic manipulation tasks, including shared autonomy, deformable-object manipulation, and precise assembly refinement~\cite{schaff2020residual,alakuijala2021residual,chi2022irp,rana2023residual,ankile2025resip}.
More broadly, skill-prior and behavior-prior methods learn reusable latent spaces or primitive libraries that guide downstream policy learning~\cite{pertsch2021accelerating,singh2021parrot,nasiriany2022maple,nasiriany2023sailor}, while recent dexterous transfer work also uses residual modules to adapt pretrained motion priors to high-DoF bimanual manipulation~\cite{li2025maniptrans}.
Different from these works, we formulate the residual component as a stabilizer that compensates for potential failure modes of the policy under unseen scenarios, improving robustness against distribution shifts and unexpected interactions.

\begin{figure*}[t]
    \centering
    \includegraphics[width=\textwidth]{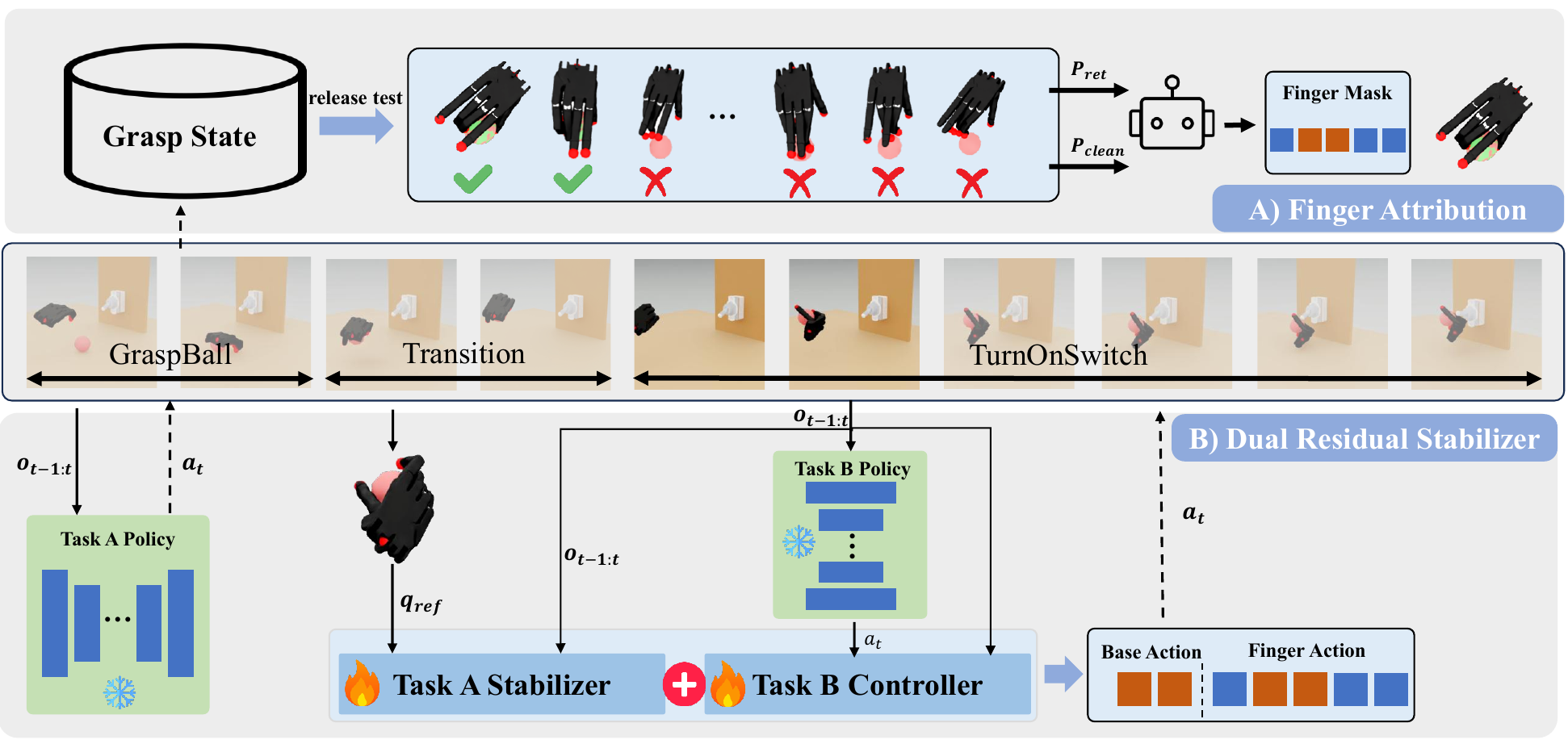}
    \vspace{-14pt}
    \caption{
    Overview of DexCompose. Given two frozen single-task policies, we first attribute fingers according to their necessity for preserving the first task and their availability for the downstream task. The selected finger-level mask is then expanded into action-space masks. During execution, a dual-residual stabilizer preserves the first task on its assigned fingers while adapting the downstream policy on the wrist and released fingers.
    }
    \label{fig:method_pipeline}
    \vspace{-4pt}
\end{figure*}

% \vspace{-4mm}
\section{Method}
\label{sec:method}
% \vspace{-2mm}
In this section, we introduce the framework of DexCompose.
We first formulate the problem of reusing pretrained policies for composed manipulation tasks (Sec.~\ref{sec:problem_formulation}). We then decompose the problem into two subproblems. The first is to discover and allocate finger-level ownership masks that determine which fingers can be reassigned to the downstream task while preserving the outcome established by the pretrained policy (Sec.~\ref{sec:agent_based_finger_allocation}). After obtaining the allocation mask, we compose the two tasks using a dual-residual stabilization framework, where one residual module maintains stability for the preserved fingers and the other adapts the downstream policy within its assigned action subspace (Sec.~\ref{sec:masked_execution_residual}). Fig.~\ref{fig:method_pipeline} summarizes the overall pipeline.

\subsection{Problem Formulation}
\label{sec:problem_formulation}

We consider sequential dexterous manipulation with two pretrained diffusion policies, $\pi_A$ and $\pi_B$. Policy $\pi_A$ executes an initial skill whose physical outcome must be preserved, such as retaining a grasped object, while policy $\pi_B$ executes a downstream interaction skill. The goal is to perform Task B while maintaining the outcome established by Task A.

At time step $t$, let $o_t$ denote the policy observation and let the action be $a_t = (p_t, r_t, q_t) \in \mathbb{R}^{d}$, where $p_t \in \mathbb{R}^{3}$ is the wrist position, $r_t \in \mathbb{R}^{3}$ denotes the wrist rotation represented in euler angles,
%\zw{check the rotation repr. type}
and $q_t \in \mathbb{R}^{d_q}$ denotes the hand joint values. Thus, the action dimension is $d = 3 + 3 + d_q$.

The two frozen policies output actions in the same action space, i.e., $a_t^A = \pi_A(o_t)$ and $a_t^B = \pi_B(o_t)$. We keep both pretrained policies fixed and learn lightweight composition models parameterized by $\Theta$ to produce the composed action
\begin{equation}
\small
    a_t^{\mathrm{comp}} = F_{\Theta}(o_t, a_t^A, a_t^B).
\end{equation}

A composed rollout is initialized from a successful Task-A state $s_0$ and follows
\begin{equation}
\small
    \tau = (s_0, a_0^{\mathrm{comp}}, s_1, a_1^{\mathrm{comp}}, \ldots, s_H).
\end{equation}
Let $S_A(\tau)$ indicate whether the outcome of Task A is preserved throughout the rollout, and $S_B(\tau)$ indicate whether Task B succeeds. The objective of policy composition is then
\begin{equation}
\small
    \max_{\Theta}
    \;
    \mathbb{E}_{\tau \sim \pi_{\mathrm{comp},\Theta}}
    \left[
        S_A(\tau) S_B(\tau)
    \right].
\end{equation}

\subsection{Finger Attribution}
\label{sec:agent_based_finger_allocation}
To reduce interference between the two policies, we assign a subset of fingers to preserve the Task-A grasp while releasing the remaining fingers for Task B.
Let $\mathcal{F}=\{1,\dots,F\}$ denote the finger set ($F=5$), and let
$m \in \{0,1\}^F$
be a binary ownership mask, where $m_f=1$ indicates that finger $f$ is preserved for Task A and $m_f=0$ indicates that it is released for Task B.
The finger-level mask is expanded to a joint-level mask $\bar m \in \{0,1\}^{d_q}$ over the hand joints. During policy composition, Task A only controls the preserved finger joints, while Task B controls the released finger joints together with the wrist DoFs.

To identify releasable fingers, we collect successful post-grasp states after Task A execution.
\begin{equation}
\small
    \mathcal D_{\mathrm{hold}}
    =
    \{(s_i^{\mathrm{hold}}, q_i^{\mathrm{ref}})\}_{i=1}^N.
\end{equation}

For each candidate mask $m$, preserved fingers replay the reference grasp while released fingers are gradually opened:
\begin{equation}
\small
    q_{i,t}^{\mathrm{test}}
    =
    \bar m \odot q_i^{\mathrm{ref}}
    +
    (1-\bar m)
    \odot
    \left[
        (1-\alpha_t) q_i^{\mathrm{ref}}
        +
        \alpha_t q^{\mathrm{open}}
    \right].
\end{equation}
We evaluate each mask using two metrics: the retention rate $P_{\mathrm{ret}}(m)$, which measures whether Task A remains successful throughout the release test, and the clean-release rate $P_{\mathrm{clean}}(m)$, which measures whether released fingers disengage without sustained contact support.
To balance retention stability, clean release quality, and residual dexterity, the final mask is selected using a task-aware allocation agent:
\begin{equation}
\small
    m^*
    =
    \operatorname{AgentSelect}
    \left(
        \{
        m,
        P_{\mathrm{ret}}(m),
        P_{\mathrm{clean}}(m),
        \eta_B(m)
        \}_{m\in\mathcal M}
    \right)
\end{equation}
where $\eta_B(m)$ denotes the residual dexterity made available by mask $m$. 

\subsection{Dual Residual Stabilizer}
\label{sec:masked_execution_residual}

After selecting the finger attribution mask $m^*$, we convert it to joint-level mask $\bar{m}^*$. These masks define action ownership during execution. Task A can only affect the preserved-finger joints, while Task B controls the wrist and the released-finger joints.
At the beginning of each composed rollout, after Task A has succeeded, we store the Task-A hand reference $q^{\mathrm{ref}}$. This reference provides the nominal preserved-finger configuration. To compensate for disturbances caused by downstream motion and interaction, we learn a bounded Task-A residual:
\begin{equation}
\small
    \Delta q_t^A
    =
    \beta_A
    \odot
    \tanh
    \left(
        \pi_{\mathrm{res}}^A(o_t, m^*)
    \right),
    \label{eq:task_a_residual}
\end{equation}
where $\pi_{\mathrm{res}}^A$ is the Task-A residual policy and $\beta_A \in \mathbb{R}_{+}^{d_q}$ is a per-joint residual bound. The Task-A preservation command is
\begin{equation}
\small
    q_t^{A,\mathrm{pres}}
    =
    q^{\mathrm{ref}}
    +
    \bar{m}^*
    \odot
    \Delta q_t^A.
    \label{eq:task_a_preserve_q}
\end{equation}
We write the corresponding full action vector as
\begin{equation}
\small
    a_t^{A,\mathrm{pres}}
    =
    \left(
        \mathbf{0}_{6},
        q_t^{A,\mathrm{pres}}
    \right).
    \label{eq:task_a_preserve_action}
\end{equation}
The zero wrist entries are ignored by the final action mask because wrist control is assigned to Task B.
For Task-B execution, the frozen downstream policy produces a nominal action $a_t^B = \pi_B(o_t)$.

We then learn a bounded residual correction in the Task-B-owned action subspace:
\begin{equation}
\small
    \Delta a_t^B
    =
    M_B^*
    \odot
    \left[
        \beta_B
        \odot
        \tanh
        \left(
            \pi_{\mathrm{res}}^B(o_t, a_t^B, m^*)
        \right)
    \right],
    \label{eq:task_b_residual}
\end{equation}
where $\pi_{\mathrm{res}}^B$ is the Task-B residual policy and $\beta_B \in \mathbb{R}_{+}^{d}$ is a per-action-dimension residual bound. The Task-B execution command is
\begin{equation}
\small
    a_t^{B,\mathrm{exec}}
    =
    a_t^B
    +
    \Delta a_t^B.
    \label{eq:task_b_exec_action}
\end{equation}
The final composed action is assembled by the fixed ownership masks:
\begin{equation}
\small
    a_t^{\mathrm{comp}}
    =
    M_A^*
    \odot
    a_t^{A,\mathrm{pres}}
    +
    M_B^*
    \odot
    a_t^{B,\mathrm{exec}}.
    \label{eq:final_composed_action}
\end{equation}
Both residual modules are initialized near zero. Therefore, before residual training, the composed policy executes the stored Task-A reference on preserved fingers and the frozen Task-B policy on the wrist and released fingers. During training, $m^*$, $\pi_A$, and $\pi_B$ remain fixed, and only $\pi_{\mathrm{res}}^A$ and $\pi_{\mathrm{res}}^B$ are optimized.
%===============================================================================

%===============================================================================
% \vspace{-4mm}
\section{Experiments}
\label{sec:experiments}
% \vspace{-2mm}

\begin{table*}[t]
\footnotesize
\setlength{\tabcolsep}{3.5pt}
\centering
\caption{
\textbf{Main comparison on 16 composite tasks.}
Composite success rate (\%), reported as mean $\pm$ standard deviation over eight seeds.
``Decomp.'' denotes Decomposed Action Space, ``FullRes.'' denotes Residual Learning, and ``Ours-ZS'' denotes the zero-shot DexCompose variant without the Task-B Residual.
}
\vspace{-6pt}
\begin{tabular}{l l c c c c c}
\toprule
& & \multicolumn{5}{c}{Composite Success Rate (\%)} \\
\cmidrule(lr){3-7}
First Task & Second Task
& Frozen & Decomp. & FullRes.
& Ours-ZS & Ours \\
\midrule

\multirow{4}{*}{GraspBall}
& OpenDoor      & $0.00_{\pm 0.00}$ & $38.94_{\pm 2.43}$ & $66.35_{\pm 2.58}$ & $73.56_{\pm 6.94}$ & \textbf{82.69}$_{\pm 8.33}$ \\
& PushButton    & $7.69_{\pm 1.92}$ & $42.31_{\pm 2.16}$ & $74.04_{\pm 2.11}$ & $76.92_{\pm 4.69}$ & \textbf{85.58}$_{\pm 7.50}$ \\
& OpenMicrowave & $0.00_{\pm 0.00}$ & $34.62_{\pm 2.57}$ & $63.46_{\pm 2.73}$ & $64.66_{\pm 7.70}$ & \textbf{73.08}$_{\pm 9.25}$ \\
& TurnOnSwitch    & $0.00_{\pm 0.00}$ & $31.73_{\pm 2.68}$ & $55.29_{\pm 3.08}$ & $63.94_{\pm 8.33}$ & \textbf{71.88}$_{\pm 8.93}$ \\

\addlinespace

\multirow{4}{*}{PourMug}
& OpenDoor      & $0.00_{\pm 0.00}$ & $36.54_{\pm 2.47}$ & $58.65_{\pm 2.81}$ & $66.59_{\pm 8.78}$ & \textbf{75.72}$_{\pm 8.08}$ \\
& PushButton    & $9.13_{\pm 2.11}$ & $40.87_{\pm 2.24}$ & $71.63_{\pm 2.17}$ & $78.61_{\pm 6.61}$ & \textbf{85.58}$_{\pm 7.28}$ \\
& OpenMicrowave & $0.00_{\pm 0.00}$ & $32.21_{\pm 2.61}$ & $54.81_{\pm 2.94}$ & $52.40_{\pm 7.67}$ & \textbf{62.74}$_{\pm 8.21}$ \\
& TurnOnSwitch    & $0.00_{\pm 0.00}$ & $30.29_{\pm 2.72}$ & $50.96_{\pm 3.16}$ & $55.29_{\pm 7.48}$ & \textbf{64.90}$_{\pm 9.89}$ \\

\addlinespace

\multirow{4}{*}{PickCan}
& OpenDoor      & $0.00_{\pm 0.00}$ & $35.58_{\pm 2.51}$ & $56.73_{\pm 2.84}$ & $72.84_{\pm 6.68}$ & \textbf{79.09}$_{\pm 6.04}$ \\
& PushButton    & $8.65_{\pm 2.08}$ & $39.90_{\pm 2.27}$ & $66.83_{\pm 2.46}$ & $77.64_{\pm 5.08}$ & \textbf{85.34}$_{\pm 6.92}$ \\
& OpenMicrowave & $0.00_{\pm 0.00}$ & $31.25_{\pm 2.64}$ & $53.85_{\pm 2.88}$ & $64.42_{\pm 11.09}$ & \textbf{72.36}$_{\pm 11.82}$ \\
& TurnOnSwitch    & $0.00_{\pm 0.00}$ & $29.81_{\pm 2.74}$ & $55.29_{\pm 2.67}$ & $58.41_{\pm 6.55}$ & \textbf{65.38}$_{\pm 6.81}$ \\

\addlinespace

\multirow{4}{*}{PickStick}
& OpenDoor      & $0.00_{\pm 0.00}$ & $83.17_{\pm 1.44}$ & $71.15_{\pm 2.17}$ & $78.37_{\pm 5.63}$ & \textbf{86.30}$_{\pm 5.78}$ \\
& PushButton    & $21.15_{\pm 3.27}$ & $84.13_{\pm 1.31}$ & $76.44_{\pm 1.94}$ & $79.57_{\pm 5.79}$ & \textbf{86.54}$_{\pm 4.84}$ \\
& OpenMicrowave & $0.00_{\pm 0.00}$ & $76.92_{\pm 1.86}$ & $61.54_{\pm 2.84}$ & $77.88_{\pm 6.69}$ & \textbf{87.26}$_{\pm 7.81}$ \\
& TurnOnSwitch    & $2.88_{\pm 1.44}$ & $61.54_{\pm 2.39}$ & $48.56_{\pm 3.22}$ & $66.59_{\pm 6.17}$ & \textbf{74.52}$_{\pm 6.10}$ \\

\cmidrule(lr){2-7}
\textbf{Mean} &
& $3.09_{\pm 0.72}$
& $45.61_{\pm 1.94}$
& $61.60_{\pm 2.31}$
& $69.23_{\pm 1.60}$
& \textbf{77.43}$_{\pm 2.64}$ \\
\bottomrule
\end{tabular}
\label{tab:main_results}
\vspace{-12pt}
\end{table*}

We evaluate on 16 composite dexterous manipulation tasks.
We first describe the experimental setup in Sec.~\ref{sec:experimental_setup} and then compare with representative policy-composition methods in Sec.~\ref{sec:main_results}, demonstrating that finger attribution and dual-residual stabilizer consistently reuse the base policies and improve composite success.
Sec.~\ref{sec:ablation_analysis} presents a series of diagnostic and ablation studies, including base-policy preservation analysis, failure-mode analysis, and component ablations, to better understand the role of each module in successful policy composition.

\subsection{Experimental Setup}
\label{sec:experimental_setup}

\noindent \textbf{Simulation Environments.}
All simulation experiments are conducted in Isaac Lab~\cite{isaaclab2024} using a Shadow Hand~\cite{shadowhand}. We instantiate Task~A with four object-retention skills:
GraspBall, PourMug, PickCan, and PickStick.
We instantiate Task~B with four interaction skills:
OpenDoor, PushButton, OpenMicrowave, and TurnOnSwitch.
This produces $4 \times 4 = 16$ composite task combinations.
All task combinations follow the same evaluation protocol.

\noindent \textbf{Datasets.}
For each base task, we collect 50 human demonstrations to train the corresponding base policy.
During Dual Stabilizer training, we additionally collect 4096 held states for each Task~A skill to support stabilization and cross-task composition.

\noindent \textbf{Evaluation Metrics.}
We primarily evaluate performance using the \emph{Composition Success Rate}, which measures whether both tasks in a composed behavior are successfully completed within a rollout. A rollout is considered successful only if the Task~A object remains retained throughout execution and the Task~B interaction task succeeds, i.e., $S_{\mathrm{comp}} = S_A \land S_B$, where $S_A$ denotes Task~A preservation success and $S_B$ denotes Task~B success. For each task composition, we evaluate $32 \times 8$ rollouts, corresponding to 32 rollouts over 8 random seeds, and report the average success rate across seeds.

\subsection{Comparison with Baselines}
\label{sec:main_results}

We compare DexCompose against four baselines that represent different strategies for policy composition.
\textbf{Frozen grasp} sequentially executes Task A and Task B, freezing the fingers during Task B to preserve the grasp established in Task A.
\textbf{Decomposed Action Space} composes the two policies by decomposing the overall action space into predefined subspaces.
\textbf{Residual Learning} augments the combined policy with a learned residual correction.
\textbf{Ours-ZS} introduces finger allocation together with the Task~A stabilizer while directly applying the frozen Task~B policy.
\textbf{Ours} combines finger attribution with dual residual stabilization by learning a Task~B residual under the finger assignment. 
Detailed baseline implementations are provided in the Appendix~\ref{sec:appendix_baselines}.

Table~\ref{tab:main_results} reports composite success for all methods.
% \my{wrong table here?}
Frozen grasp fails almost completely, indicating that directly linking two independently trained policies introduces severe interference between their policy distributions during execution.
Both the decomposed baseline and the residual learning method substantially improve over direct chaining. However, their performance remains unstable across different task combinations and still performs poorly on interaction-heavy tasks.
Our full method achieves the best performance across all task combinations, reaching $77.4\%$ mean composite success and outperforming the strongest baseline by $15.8$ percentage points, with the largest gains observed on challenging tasks such as OpenMicrowave and TurnOnSwitch. These consistent improvements across all task families demonstrate the effectiveness of our DexCompose framework in maintaining stable multi-task manipulation during sequential policy execution.

\begin{figure*}[t]
    \centering
    \includegraphics[width=\textwidth]{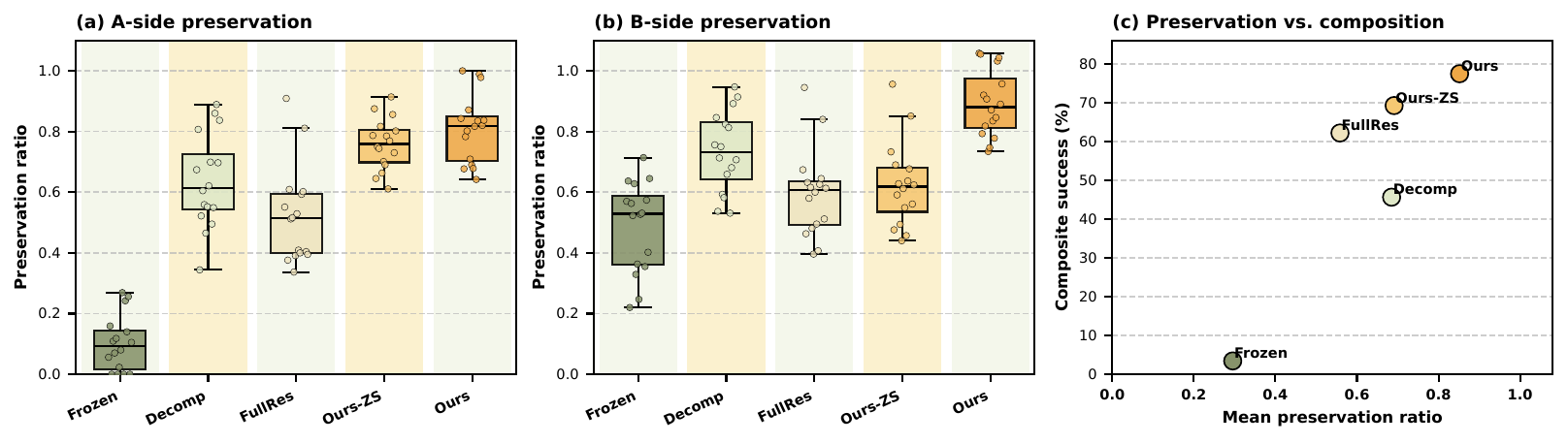}
    \vspace{-18pt}
    \caption{
    Base-policy preservation analysis.
    We measure A-side and B-side preservation ratios to evaluate whether composite policies retain the capabilities of the original base policies.
    Frozen-Grasp preserves part of the Task-B behavior but fails to maintain Task A, while Ours achieves high preservation on both sides.
    }
    \vspace{-7mm}
    \label{fig:base_policy_preservation}
\end{figure*}

\subsection{Ablation and Diagnostic Analysis}
\label{sec:ablation_analysis}

We conduct studies to isolate the contribution of each component and understand failure modes. 

\begin{wrapfigure}{r}{0.52\linewidth}
    \centering
    \vspace{-4mm}
    \includegraphics[width=0.5\textwidth]{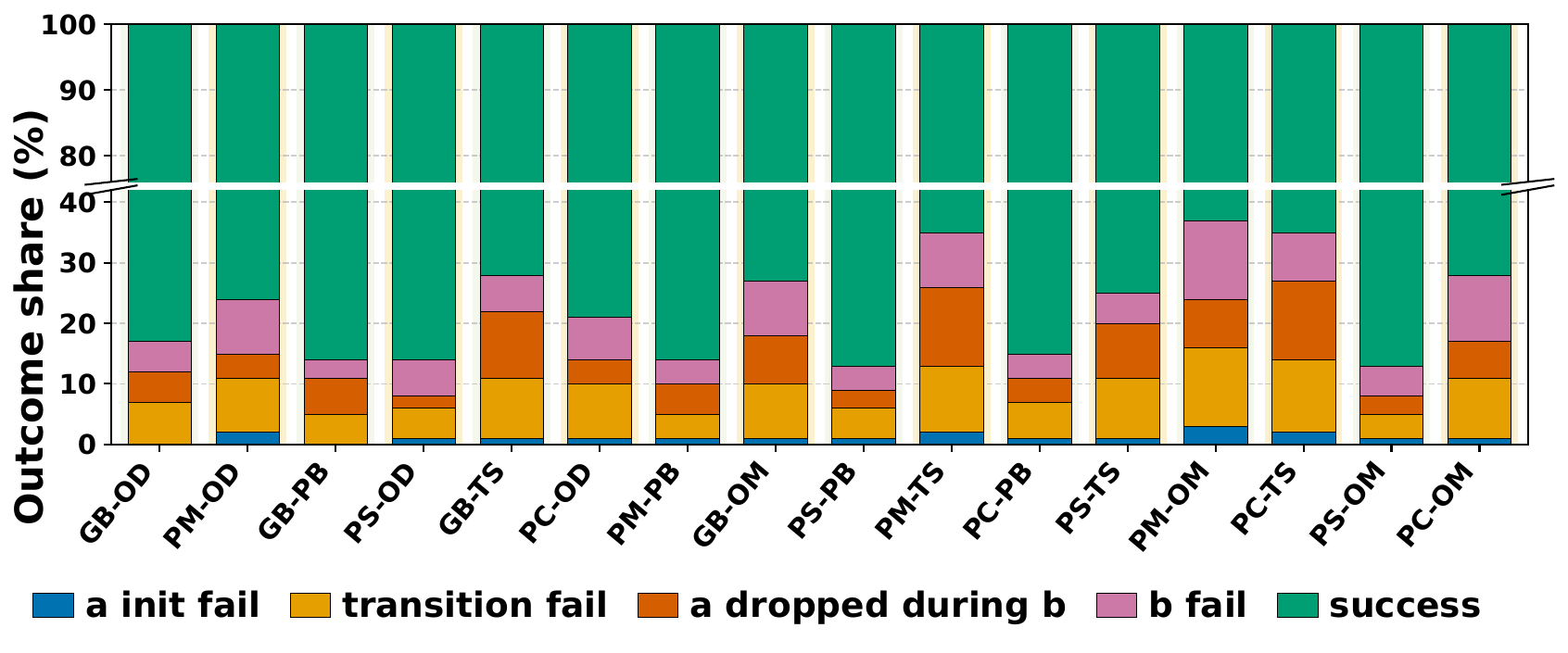}
    \vspace{-8pt}
    \caption{
    Failure-mode breakdown across task combinations.
    }
    \label{fig:failure_breakdown}
    \vspace{-4mm}
\end{wrapfigure}

\noindent \textbf{Base-policy preservation.}
% \paragraph{Base-policy preservation.}
A key point in policy composition is that combining multiple tasks should not significantly degrade the performance of the original base policies. Figure~\ref{fig:base_policy_preservation} evaluates this property by measuring how well each method preserves the capabilities of the original base policies during composite execution.
Frozen-Grasp obtains very low A-side preservation at $0.102$, indicating that the first skill is frequently lost either before or during execution of the second task. However, its B-side preservation remains around $0.489$, suggesting that the second-skill policy itself is not destroyed. Decomposed method shows substantial variation across compositions, performing well on easier combinations but remaining brittle when tasks compete for hand resources.
In contrast, Ours achieves A-side and B-side preservation ratios of $0.811$ and $0.893$, respectively, and occupies the upper-right region of the preservation-versus-composition plot, combining high mean preservation with the highest composite success. These results demonstrate that our method not only composes policies effectively, but also preserves the original task capabilities as much as possible.

\noindent \textbf{Failure-mode analysis.}
Across all task combinations, the distribution of failures varies significantly depending on the second task. Transition failures are the dominant failure source, indicating that the transition stage is the most challenging and that the stabilizer is critical for maintaining grasp stability. TurnOnSwitch causes more preservation failures, including object drops and \textit{B-fail} outcomes, due to grasp perturbations during switch contact. In contrast, OpenMicrowave exhibits more \textit{B-fail} cases without large increases in drops, suggesting that actuating the microwave handle under constrained grasp configurations is itself difficult.Less perturbative second tasks are comparatively stable, producing fewer preservation failures and higher overall success rates. Overall, the results show that task-specific interaction dynamics strongly influence dual-task execution robustness.

\noindent \textbf{Component ablations.}
Table~\ref{tab:ablation_study} isolates the contribution of each module. Removing the Task-A residual causes the most severe degradation, confirming it as the essential mechanism for preserving the first-task outcome.
Removing Finger Allocation and Action Masking each reduce success substantially, showing that explicit ownership assignment is critical when the two tasks compete for hand resources. 
The Action Masking result is particularly informative: training a residual that writes to all action dimensions, as in conventional residual learning~\cite{johannink2019residual,silver2018residual}, performs much worse than our masked variant, confirming that structural ownership enforcement is strongly beneficial for two-policy composition. 
The Task-B residual and transition stage have smaller but meaningful effects, primarily improving execution refinement and stage-switching stability. 
Overall, the ablation reveals a clear hierarchy: Task-A preservation is essential, finger allocation and action masking reduce interference, and the Task-B residual with the transition stage further improves robustness.

\begin{table*}[t]
\footnotesize
\setlength{\tabcolsep}{3pt}
\centering
\caption{
\textbf{Component ablation on 16 composite tasks.}
Composite success rate (\%), reported as mean $\pm$ standard deviation over eight seeds.
Ours-ZS denotes the zero-shot DexCompose variant without the Task-B Residual.
The remaining columns remove one component from the full DexCompose method:
$-$Finger removes Finger Allocation, $-$Task-A removes the Task-A Residual,
$-$Mask removes Action Masking, and $-$Trans. removes the Transition Stage.
% Bold denotes the full method.
}
\vspace{-6pt}
\begin{tabular}{l l c c c c c c}
\toprule
& & \multicolumn{6}{c}{Composite Success Rate (\%)} \\
\cmidrule(lr){3-8}
First Task & Second Task
& Ours
& Ours-ZS
& $-$Finger
& $-$Task-A
& $-$Mask
& $-$Trans. \\
\midrule

\multirow{4}{*}{GraspBall}
& OpenDoor      & \textbf{82.69}$_{\pm 8.33}$ & $73.56_{\pm 6.94}$ & $73.80_{\pm 7.16}$ & $3.85_{\pm 3.86}$ & $67.07_{\pm 8.23}$ & $75.00_{\pm 7.97}$ \\
& PushButton    & \textbf{85.58}$_{\pm 7.50}$ & $76.92_{\pm 4.69}$ & $70.43_{\pm 9.43}$ & $3.37_{\pm 3.71}$ & $68.75_{\pm 8.37}$ & $79.33_{\pm 7.61}$ \\
& OpenMicrowave & \textbf{73.08}$_{\pm 9.25}$ & $64.66_{\pm 7.70}$ & $43.51_{\pm 13.29}$ & $0.96_{\pm 1.50}$ & $56.73_{\pm 10.66}$ & $64.90_{\pm 9.73}$ \\
& TurnOnSwitch  & \textbf{71.88}$_{\pm 8.93}$ & $63.94_{\pm 8.33}$ & $54.33_{\pm 8.80}$ & $0.96_{\pm 1.97}$ & $53.37_{\pm 7.49}$ & $63.22_{\pm 7.56}$ \\

\addlinespace

\multirow{4}{*}{PourMug}
& OpenDoor      & \textbf{75.72}$_{\pm 8.08}$ & $66.59_{\pm 8.78}$ & $62.02_{\pm 8.64}$ & $15.38_{\pm 5.48}$ & $56.97_{\pm 6.65}$ & $64.90_{\pm 7.88}$ \\
& PushButton    & \textbf{85.58}$_{\pm 7.28}$ & $78.61_{\pm 6.61}$ & $62.74_{\pm 8.70}$ & $17.79_{\pm 6.55}$ & $67.55_{\pm 6.82}$ & $76.68_{\pm 6.07}$ \\
& OpenMicrowave & \textbf{62.74}$_{\pm 8.21}$ & $52.40_{\pm 7.67}$ & $37.02_{\pm 6.85}$ & $13.46_{\pm 5.62}$ & $45.43_{\pm 6.58}$ & $55.05_{\pm 7.71}$ \\
& TurnOnSwitch  & \textbf{64.90}$_{\pm 9.89}$ & $55.29_{\pm 7.48}$ & $40.62_{\pm 9.29}$ & $12.02_{\pm 5.54}$ & $44.23_{\pm 7.09}$ & $55.77_{\pm 8.54}$ \\

\addlinespace

\multirow{4}{*}{PickCan}
& OpenDoor      & \textbf{79.09}$_{\pm 6.04}$ & $72.84_{\pm 6.68}$ & $57.21_{\pm 11.30}$ & $14.66_{\pm 5.32}$ & $60.82_{\pm 8.43}$ & $70.43_{\pm 6.82}$ \\
& PushButton    & \textbf{85.34}$_{\pm 6.92}$ & $77.64_{\pm 5.08}$ & $56.49_{\pm 10.56}$ & $16.59_{\pm 6.55}$ & $62.26_{\pm 10.48}$ & $77.64_{\pm 7.20}$ \\
& OpenMicrowave & \textbf{72.36}$_{\pm 11.82}$ & $64.42_{\pm 11.09}$ & $30.29_{\pm 10.16}$ & $11.78_{\pm 5.87}$ & $51.68_{\pm 10.49}$ & $62.02_{\pm 9.95}$ \\
& TurnOnSwitch  & \textbf{65.38}$_{\pm 6.81}$ & $58.41_{\pm 6.55}$ & $32.21_{\pm 6.55}$ & $11.54_{\pm 4.84}$ & $45.43_{\pm 7.72}$ & $57.93_{\pm 6.20}$ \\

\addlinespace

\multirow{4}{*}{PickStick}
& OpenDoor      & \textbf{86.30}$_{\pm 5.78}$ & $78.37_{\pm 5.63}$ & $73.32_{\pm 8.52}$ & $5.53_{\pm 5.13}$ & $71.88_{\pm 6.38}$ & $77.40_{\pm 4.45}$ \\
& PushButton    & \textbf{86.54}$_{\pm 4.84}$ & $79.57_{\pm 5.79}$ & $70.91_{\pm 9.23}$ & $6.25_{\pm 4.23}$ & $68.27_{\pm 7.95}$ & $78.12_{\pm 4.94}$ \\
& OpenMicrowave & \textbf{87.26}$_{\pm 7.81}$ & $77.88_{\pm 6.69}$ & $54.33_{\pm 8.61}$ & $3.37_{\pm 2.98}$ & $68.27_{\pm 6.73}$ & $77.64_{\pm 8.45}$ \\
& TurnOnSwitch  & \textbf{74.52}$_{\pm 6.10}$ & $66.59_{\pm 6.17}$ & $53.12_{\pm 10.36}$ & $2.88_{\pm 4.13}$ & $57.45_{\pm 7.71}$ & $68.27_{\pm 7.09}$ \\

\cmidrule(lr){2-8}
\textbf{Average} & --
& \textbf{77.43}$_{\pm 2.64}$
& $69.23_{\pm 1.60}$
& $54.52_{\pm 2.09}$
& $9.13_{\pm 1.29}$
& $59.13_{\pm 1.75}$
& $69.02_{\pm 2.07}$ \\
\bottomrule
\end{tabular}
\label{tab:ablation_study}
\vspace{-12pt}
\end{table*}

\begin{wraptable}{r}{0.36\linewidth}
\vspace{-4mm}
\centering
\small
\caption{Heuristic vs. LLM agent mask selection.}
\vspace{-6pt}
\setlength{\tabcolsep}{4pt}
\begin{tabular}{lcc}
\toprule
Task Pair & Heur. & LLM \\
\midrule
GraspBall+OpenDoor   & 72.1 & \textbf{82.7} \\
GraspBall+Switch & 63.5 & \textbf{71.9} \\
PickCan+Micro.   & 68.3 & \textbf{72.4} \\
PourMug+Switch   & 62.0 & \textbf{64.9} \\
\midrule
Mean & 66.5 & \textbf{73.0} \\
\bottomrule
\end{tabular}
\label{tab:llm_vs_heuristic}
\vspace{-12pt}
\end{wraptable}

\noindent \textbf{Agent-based mask selection.}
As in Table~\ref{tab:llm_vs_heuristic}, we compare the LLM agent-based mask selector with a heuristic baseline that selects the mask with the highest object-retention rate under a minimum finger-release constraint.
%
% This local criterion can be suboptimal for long-horizon manipulation.
While effective for short-term stability, this local criterion can be suboptimal as it ignores downstream finger utility.
For example, in GraspBall, holding the ball with the thumb and index finger yields about 7\% higher retention success than a thumb–ring–little grip,
% using the thumb, ring finger, and little finger, 
so the heuristic selects the thumb–index mask. However, this occupies the index finger, limiting downstream actions such as pressing buttons or opening doors. The LLM instead chooses the lower-retention thumb–ring–little grip, keeping the index finger free for subsequent tasks.
See Appendix~\ref{sec:appendix_llm} for detailed analysis.

%===============================================================================
% \vspace{-4mm}
\section{Conclusion}
\label{sec:conclusion}
% \vspace{-2mm}
We presented DexCompose, a framework for composing pretrained dexterous manipulation policies by treating embodiment redundancy as a reusable resource. DexCompose identifies redundant action dimensions, assigns structured action ownership through finger-aware masking, and applies dual residual stabilizers within the allocated subspaces. Our results show that structured action ownership reduces cross-task interference and is critical for successful composite manipulation. More broadly, DexCompose points toward a scalable framework for reusable dexterous manipulation, enabling learned skills to be flexibly composed without requiring task-specific retraining of the base policies.

\noindent \textbf{Limitations and Future Work.}
Our current framework focuses on composing two sequential skills at a time. Extending the framework to longer-horizon compositions involving multiple interacting skills remains an important direction for future work.

%===============================================================================

% \clearpage

%===============================================================================


% \clearpage 
% The acknowledgments are automatically included only in the final and preprint versions of the paper.
% \acknowledgments{If a paper is accepted, the final camera-ready version will (and probably should) include acknowledgments. All acknowledgments go at the end of the paper, including thanks to reviewers who gave useful comments, to colleagues who contributed to the ideas, and to funding agencies and corporate sponsors that provided financial support.}

%===============================================================================

%===============================================================================
\appendix

% Appendix-only TOC (separate from the main paper)
\makeatletter
\newcommand{\apxtableofcontents}{%
  \section*{\LARGE Appendix}%
  \@starttoc{apx}% writes/reads \jobname.apx
  \vspace{0.5em}%
}

% \newcommand{\apxaddsection}[1]{%
%   \addcontentsline{apx}{section}{\protect\numberline{\thesection}#1}%
% }
% \newcommand{\apxaddsubsection}[1]{%
%   \addcontentsline{apx}{subsection}{\protect\numberline{\thesubsection}#1}%
% }

% % Hook sectioning commands *only* for the appendix part of the document.
% \let\apx@old@section\section
% \renewcommand{\section}{\@ifstar{\apx@old@section*}{\apx@section}}
% \newcommand{\apx@section}[1]{\apx@old@section{#1}\apxaddsection{#1}}

% \let\apx@old@subsection\subsection
% \renewcommand{\subsection}{\@ifstar{\apx@old@subsection*}{\apx@subsection}}
% \newcommand{\apx@subsection}[1]{\apx@old@subsection{#1}\apxaddsubsection{#1}}
% \makeatother

% Print the appendix-only TOC here
\apxtableofcontents

\section{Base Policy Training and Evaluation Protocol}

We train one diffusion-based policy for each primitive dexterous manipulation task using replayed successful human demonstrations collected in Isaac Lab. The eight base policies correspond to four object-retention skills (GraspBall, PickStick, PickCan, PourMug) and four downstream interaction skills (OpenDoor, PushButton, OpenMicrowave, TurnOnSwitch).

\noindent\textbf{Observation and action spaces.}
All policies operate on low-dimensional proprioceptive and task-state observations. The observation consists of normalized hand joint positions together with task-specific geometric features such as object-relative poses, articulated-object joint states, and palm-to-object offsets. Depending on the task, the observation dimension ranges from 32 to 37.

The action space is shared across all tasks and consists of 28-dimensional Floating Shadow Hand joint-position commands:
\[
a_t = (p_t, r_t, q_t),
\]
where \(p_t \in \mathbb{R}^3\) denotes wrist translation, \(r_t \in \mathbb{R}^3\) denotes wrist rotation, and \(q_t\) denotes the 22 finger-joint position targets of the Shadow Hand.

\noindent\textbf{Diffusion policy architecture.}
All base policies share the same conditional diffusion architecture. Given the latest two observations (\(obs\_horizon=2\)), the policy predicts a horizon of 32 future actions (\(action\_horizon=32\)).

We use a FiLM-conditioned 1D UNet diffusion backbone. The observation history is flattened and encoded with an MLP:
\[
\text{Linear} \rightarrow \text{Mish} \rightarrow \text{Linear},
\]
and injected into the diffusion timestep embedding through FiLM-style conditioning. The denoising backbone uses channel widths \((256, 512, 1024)\) with residual temporal blocks and Mish activations.

Diffusion training uses a DDPM objective with 100 diffusion timesteps and the \texttt{squaredcos\_cap\_v2} noise schedule.

\noindent\textbf{Training details.}
Policies are trained with behavior cloning on replayed successful demonstrations. Replay trajectories are segmented into overlapping observation-action windows of shape:
\[
obs \in \mathbb{R}^{2 \times d_{obs}}, \qquad
action \in \mathbb{R}^{32 \times 28}.
\]

All policies use AdamW optimization with batch size 256, learning rate \(1\times10^{-4}\), 2000-step linear warmup followed by cosine decay, EMA checkpoint averaging, and gradient clipping with global norm 1.0.

\noindent\textbf{Inference protocol.}
At evaluation time, the policy samples an action chunk by iterative denoising from Gaussian noise. We execute only the first several actions from the predicted sequence and then replan in a receding-horizon manner until task success or timeout.

\noindent\textbf{Standalone base-policy performance.}
Table~\ref{tab:base_policy_success} reports the standalone success rate of all pretrained base policies before policy composition.

\begin{table}[h]
\centering
\caption{Standalone success rate (\%) of pretrained base policies. Results are evaluated using the corresponding execute horizon for each task.}
\label{tab:base_policy_success}
\begin{tabular}{lcc}
\toprule
Task & Execute Horizon & Success Rate (\%) \\
\midrule
GraspBall & 8  & 100.0 \\
PickStick & 16 & 100.0 \\
PickCan   & 16 & 100.0 \\
PourMug        & 16 & 100.0 \\
OpenDoor       & 24 & 97.59 \\
OpenMicrowave  & 24 & 89.90 \\
PushButton     & 16 & 100.0 \\
TurnOnSwitch   & 30 & 82.21 \\
\bottomrule
\end{tabular}
\end{table}

\noindent\textbf{Task success criteria.}
For object-retention tasks, success is defined using task-specific geometric conditions. GraspBall succeeds when the object is within a threshold distance of the target pose. PickStick, PickCan, and PourMug additionally require the object to be lifted above a minimum height while satisfying task-specific orientation constraints.

For downstream interaction tasks, OpenDoor and OpenMicrowave succeed when the articulated joint angle exceeds a predefined opening threshold. PushButton succeeds when the button displacement exceeds a target travel distance. TurnOnSwitch succeeds when the switch joint reaches at least \(80\%\) of its reachable joint range.

For composite-task evaluation, a rollout is considered successful only if the Task-A object remains retained throughout execution and the downstream Task-B interaction succeeds:
\[
S_{\mathrm{comp}} = S_A \land S_B.
\]
All methods and baselines are evaluated under the same success definitions and rollout horizons.

\section{Learning Procedure and Implementation Details}
\label{sec:appendix}

This appendix provides the complete learning procedure for all trained components of DexCompose, addressing the reproducibility requirements for the allocation mechanism (Section~\ref{sec:agent_based_finger_allocation}) and the two residual policies (Section~\ref{sec:masked_execution_residual}). All experiments are conducted on a single NVIDIA RTX 4090 GPU.

\subsection{PPO Training: Task A Residual Stabilizer}
\label{sec:appendix_taska_ppo}

The Task~A residual policy $\pi^A_{\mathrm{res}}$ is trained via Proximal Policy Optimization (PPO)~\cite{schulman2017proximal} to produce bounded joint-space corrections that stabilize the held object against disturbances from base motion and finger release.

\noindent\textbf{Observation space.} The observation includes: preserved-finger joint positions and velocities, the previous residual output, object pose and velocity in the palm frame, fingertip-to-object distance errors, binary contact indicators, torque features at the preserved finger joints, the executed base action, and a rollout phase indicator.

\noindent\textbf{Action space.} The residual is a Gaussian policy whose mean head is initialized to zero and whose log-standard-deviation is initialized to $-1.5$. The output is bounded via $\Delta^A_t = b_A \odot \tanh(\pi^A_{\mathrm{res}}(o_t, m^*))$, where $b_A$ is a per-dimension bound vector. Zero-mean initialization ensures that the policy begins at the stored hold reference with no correction.

\noindent\textbf{Reward function.} The reward at each timestep penalizes the Euclidean distance between the held object and a target position near the palm center, plus a small penalty on the magnitude of the residual correction to discourage unnecessary adjustments:
\begin{equation}
R^A_t = -\|\mathbf{p}_{\mathrm{obj}} - \mathbf{p}_{\mathrm{palm}}\|_2 - \alpha \|\Delta^A_t\|^2_2,
\end{equation}
with $\alpha = 0.001$.

\noindent\textbf{Architecture and hyperparameters.} The actor and critic share an MLP backbone of $[256, 256, 128]$ units with ELU activations and Prefix Layer Normalization before each hidden layer. Complete hyperparameter settings are provided in Appendix~\ref{sec:appendix_hyperparams}.

\noindent\textbf{Training budget.} Training uses 1024 parallel environments with rollout length 24, for 1000 PPO iterations, yielding $24{,}576{,}000$ total environment steps. Wall-clock time is approximately 20~minutes on the RTX~4090.

\subsection{PPO Training: Task B Residual}
\label{sec:appendix_taskb_ppo}

The Task~B residual policy $\pi^B_{\mathrm{res}}$ is trained via PPO with the same backbone architecture and hyperparameters as $\pi^A_{\mathrm{res}}$ (see Appendix~\ref{sec:appendix_hyperparams}), with the following differences.

\noindent\textbf{Observation space.} In addition to the composite state, which includes Task~A held object and preserved finger states, the policy observes the nominal Task~B policy output $a^0_{B,t} = \pi_B(s_t)$ and the selected mask $m^*$. This allows the residual to condition its correction on what the frozen Task~B policy would have executed and which action dimensions are available.

\noindent\textbf{Action space.} The residual $\delta^B_t$ is produced by the same Gaussian actor architecture, bounded, and masked by $M_B$ before execution (as in Eqs.~(10)--(12)). Only action dimensions assigned to Task~B are affected.

\noindent\textbf{Reward function.} The reward combines Task~B progress and Task~A preservation:
\begin{equation}
R^B_t = R_{\mathrm{taskB}}(s_t, a_t) - \beta \|\mathbf{p}_{\mathrm{obj}} - \mathbf{p}_{\mathrm{palm}}\|_2 - \alpha \|\delta^B_t\|^2_2,
\end{equation}
where $R_{\mathrm{taskB}}$ is the environment-defined Task~B success reward (e.g., door angle for OpenDoor), and the object--palm distance term is identical to that used in Phase~1. We set $\beta = 1.0$ and $\alpha = 0.001$.

\noindent\textbf{Joint Task A fine-tuning.} During Task~B residual training, the Task~A residual $\pi^A_{\mathrm{res}}$ is optionally fine-tuned jointly with a reduced learning rate of $1 \times 10^{-4}$ while the Task~B residual trains at the standard rate ($3 \times 10^{-4}$). This allows the Task~A stabilizer to adapt to the specific disturbance patterns induced by Task~B execution without catastrophic forgetting.

\noindent\textbf{Training budget.} Training uses 8 parallel environments with rollout length 200, for 1000 PPO iterations, yielding $1{,}600{,}000$ nominal environment-step slots. Due to per-episode filtering from pre-interaction termination and early success, the effective number of samples per PPO update varies between approximately 408 and 750. Wall-clock time is approximately 4~hours on the RTX~4090.

\subsection{Training Curriculum}
\label{sec:appendix_curriculum}

The full training curriculum consists of three sequential stages:

\begin{enumerate}
    \item \textbf{Held-state collection (no learning):} For each Task~A skill (GraspBall, PourMug, PickCan, PickStick), we collect $N = 4096$ successful held-object states by rolling out the pretrained Task~A policy to completion. Each state stores the simulator state, object pose, hand configuration, contact information, and reference hold action. States are sampled uniformly during subsequent release tests.

    \item \textbf{Phase 1 (Task A residual training):} For each Task~A skill, $\pi^A_{\mathrm{res}}$ is trained independently (not yet coupled to a specific Task~B) using the reward defined in Appendix~\ref{sec:appendix_taska_ppo}. The policy learns to stabilize the held object against randomized base and finger perturbations. After training, $\pi^A_{\mathrm{res}}$ is frozen.

    \item \textbf{Phase 2 (Finger allocation and Task B residual training):} For each composite task pair:
    \begin{enumerate}
        \item The LLM selects a finger mask $m^*$ from the candidate set $\mathcal{M}$ using release-test diagnostics computed from the stored held-state library (Appendix~\ref{sec:appendix_llm}).
        \item \textbf{Release-test evaluation:} For each candidate mask, we run $K=100$ release-test rollouts. Each rollout restores a held state sampled uniformly from the library and runs the masked replay protocol for $T_{\mathrm{test}}$ steps (Appendix~\ref{sec:appendix_release_test}).
        \item With $m^*$ fixed, $\pi^B_{\mathrm{res}}$ is trained with $\pi^A_{\mathrm{res}}$ and $\pi_B$ frozen (optionally with joint Task~A fine-tuning, Appendix~\ref{sec:appendix_taskb_ppo}).
    \end{enumerate}
\end{enumerate}

No joint fine-tuning of the full composite system is performed beyond the optional joint Task~A update during Phase~2. The frozen Task~B policy $\pi_B$ is never updated.

\subsection{Release-Test Protocol}
\label{sec:appendix_release_test}

For each candidate mask $m \in \mathcal{M}$, we perform the following release-test procedure:

\begin{enumerate}
    \item \textbf{Restore held state:} The simulator is reset to a held state $s_A^j$ sampled uniformly from $\mathcal{D}_{\mathrm{hold}}$.
    \item \textbf{Masked replay:} Fingers in $G_A(m)$ replay the stored Task~A hold reference action. Fingers in $G_B(m)$ are driven toward an open or neutral pose following the linear ramp schedule:
    \begin{equation}
    \lambda_t = \frac{t}{T_{\mathrm{test}}},
    \end{equation}
    where $T_{\mathrm{test}} = 100$ simulation steps. At $t = 0$, released fingers are at the hold configuration. At $t = T_{\mathrm{test}}$, they reach the fully open pose.
    \item \textbf{Retention check:} The object is \textit{retained} if it remains within a distance threshold $d_{\mathrm{ret}}$ of the palm center and above a drop-height threshold $h_{\mathrm{drop}}$ at the end of the test.
    \item \textbf{Clean-release check:} Released fingers are considered \textit{free} if their contact forces fall below a threshold $F_{\mathrm{min}}$ during the final 20 steps of the test, indicating they no longer provide sustained support.
    \item \textbf{Aggregation:} We repeat Steps 1--4 for $K$ independent rollouts (default $K=100$) per candidate mask by resampling held states from $\mathcal{D}_{\mathrm{hold}}$. $P_{\mathrm{ret}}(m)$ and $P_{\mathrm{clean}}(m)$ are computed as the fraction of rollouts passing each check (as defined in the main text).
\end{enumerate}

We use $d_{\mathrm{ret}} = 0.05$~m, $h_{\mathrm{drop}} = 0.03$~m, and $F_{\mathrm{min}} = 0.1$~N.

\subsection{Baseline Implementation Details}
\label{sec:appendix_baselines}

We provide detailed implementation descriptions for the baselines compared in Section~4.2.

\noindent\textbf{Frozen grasp (``Frozen'' in Table~\ref{tab:main_results}).}
This baseline sequentially executes the frozen Task~A policy $\pi_A$ and frozen Task~B policy $\pi_B$. After Task~A success is detected, we execute $\pi_B$ while \emph{freezing the grasp-maintaining finger joints} at the final Task~A configuration (i.e., those finger action dimensions are held constant during Task~B). All non-finger action dimensions (e.g., base and arm) are controlled by $\pi_B$. This baseline therefore preserves the grasp by preventing Task~B from writing to the frozen finger dimensions, but it does not perform finger allocation, action masking for released fingers, release-test validation, or residual correction.

\noindent\textbf{Decomposed Action Space.}
We retrain both base policies with an auxiliary loss that encourages each policy to minimize action magnitudes on dimensions assigned to the other task. Specifically, for a fixed finger allocation mask $m$, the Task~A policy is trained with an additional penalty $\lambda \|M_B \odot a^A_t\|^2_2$ on actions in the Task~B subspace, and symmetrically for the Task~B policy. The allocation mask for this baseline is fixed per task family. For GraspBall and PourMug, thumb, index, and middle are preserved. For PickCan, index, middle, and ring are preserved. For PickStick, thumb and index are preserved. These assignments are chosen based on the dominant contact pattern observed in single-task rollouts and are not optimized per task pair. The penalty coefficient $\lambda = 0.1$ is tuned on a held-out validation pair. Both policies are retrained for the same number of environment steps as the original single-task policies. During composite execution, actions from both policies are combined via the same hard-mask overwrite as our method (Eq.~(12)), but neither policy receives residual corrections or release-test validation.

\noindent\textbf{Residual Learning.}
We adapt the residual RL formulation of Johannink et al.~\cite{johannink2019residual} to the two-stage composition setting. A single Q-function and policy are trained to output corrections to the combined nominal action:
\begin{equation}
a_t = \underbrace{M_A \odot a^{\mathrm{ref}}_A + M_B \odot \pi_B(s_t)}_{\text{nominal}} + \delta_t,
\end{equation}
where $\delta_t \in \mathbb{R}^d$ is an unconstrained residual (no action masking applied to the correction), $a^{\mathrm{ref}}_A$ is the stored Task~A hold action, and $M_A, M_B$ use the same fixed allocation masks as the Decomposed baseline. The residual policy observes the full composite state and is trained via soft actor-critic with the same composite reward as our Task~B residual (Appendix~\ref{sec:appendix_taskb_ppo}). The critical difference from our method is that $\delta_t$ is not masked before execution, so the residual can write to any action dimension, including those assigned to Task~A preservation. This baseline therefore isolates the effect of structural action ownership enforcement: it uses residuals but does not enforce the residual mask ($\delta_t = M_B \odot \delta_t$), equivalently ($M_A \odot \delta_t = 0$).

\noindent \textbf{Ours-ZS.}
This ablation uses our full finger allocation pipeline with LLM-based mask selection and release-test validation, and the trained Task~A residual stabilizer, but executes the frozen Task~B policy directly without residual correction. The Task~B nominal action is masked as $M_B \odot \pi_B(s_t)$ and applied to the Task~B-owned dimensions. This isolates the contribution of the Task~B residual in adapting the frozen policy to the constrained hand state.

\begin{center}
\begin{minipage}{\linewidth}
    \centering
    \includegraphics[width=\linewidth,height=0.33\textheight,keepaspectratio]{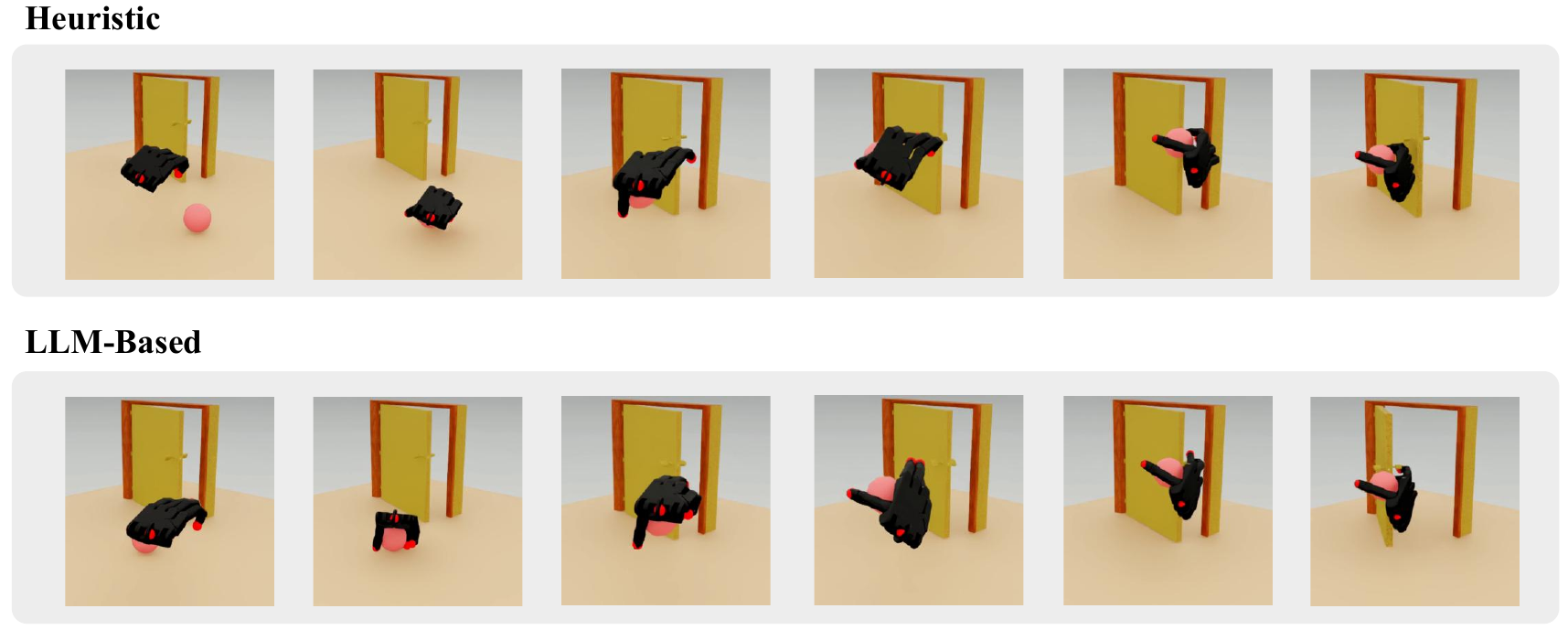}
    \captionof{figure}{
    Comparison between heuristic and LLM-based mask selection on the GraspBall+OpenDoor task.
    The heuristic chooses a thumb--index grasp because it maximizes immediate object retention, but the occupied index finger prevents effective door interaction in the subsequent stage.
    In contrast, the LLM selects a thumb--ring--little grasp that preserves the index finger for future manipulation.
    }
    \label{fig:failure_llm}
\end{minipage}
\end{center}

\subsection{Hyperparameters and Compute Resources}
\label{sec:appendix_hyperparams}

We provide the complete hyperparameter settings for reproducibility. All experiments are conducted on a single NVIDIA RTX 4090 with 24~GB of memory.

\noindent\textbf{PPO hyperparameters.}
Both Task~A and Task~B residual policies are trained with Proximal Policy Optimization using the same core settings. The learning rate is $3 \times 10^{-4}$ for standard training and reduced to $1 \times 10^{-4}$ for joint Task~A fine-tuning during Phase~2. The PPO clipping parameter $\epsilon$ is set to $0.2$, and we use GAE with $\lambda = 0.95$ and a discount factor $\gamma = 0.99$. Each PPO iteration performs $5$ update epochs over $4$ minibatches. The entropy coefficient is $0.001$, the value loss coefficient is $1.0$, and gradients are clipped to a maximum norm of $1.0$.

\noindent\textbf{Architecture.}
The actor and critic share an MLP backbone with hidden layers of sizes $[256, 256, 128]$, using ELU activations and Prefix Layer Normalization before each hidden layer. The actor outputs a Gaussian distribution over residual corrections, with the mean head initialized to zero and the log-standard-deviation initialized to $-1.5$. Both policies use this same architecture.

\noindent\textbf{Task A training budget.}
The Task~A residual stabilizer is trained with $1024$ parallel environments, a rollout length of $24$ steps per environment, yielding a nominal batch size of $24{,}576$ per PPO iteration and a minibatch size of $6144$. Training runs for $1000$ PPO iterations, totaling $24{,}576{,}000$ environment steps. Wall-clock time is approximately 20~minutes.

\noindent\textbf{Task B training budget.}
The Task~B residual is trained with $8$ parallel environments and a rollout length of $200$ steps, giving a nominal batch size of $1600$ and a minibatch size of $400$. Due to per-episode filtering from pre-interaction termination and early success, the effective number of samples per update varies between approximately $408$ and $750$. Training runs for $1000$ PPO iterations, yielding $1{,}600{,}000$ nominal environment steps. Wall-clock time is approximately 4~hours.

\noindent\textbf{Reward parameters.}
The object-to-palm distance is weighted by $\beta = 1.0$ in the Task~B composite reward. The residual magnitude penalty uses $\alpha = 0.001$ for both Task~A and Task~B policies.

\noindent\textbf{Release-test parameters.}
We collect a library of $N = 4096$ held states per Task~A skill. For each candidate finger mask, we run $K=100$ release-test rollouts by sampling held states uniformly from this library (with replacement). Each rollout lasts $T_{\mathrm{test}} = 100$ simulation steps. The retention distance threshold is $d_{\mathrm{ret}} = 0.05$~m, the drop-height threshold is $h_{\mathrm{drop}} = 0.03$~m, and the clean-release force threshold is $F_{\mathrm{min}} = 0.1$~N.

\subsection{LLM-Based Finger Mask Selection}
\label{sec:appendix_llm}

As described in Section~\ref{sec:agent_based_finger_allocation}, finger mask selection is performed by a large language model rather than a learned allocation network. We use OpenAI GPT-5.4 (snapshot \texttt{gpt-5.4-2026-03-05}, accessed May~18, 2026) with temperature $T = 0$ to ensure deterministic mask selection.

For each task pair, the LLM receives a structured prompt consisting of three components: (1) a natural-language description of the object manipulation in Task~A and the downstream interaction required in Task~B; (2) a list of candidate masks $\mathcal{M}$ together with their release-test diagnostics, including retention rate $P_{\mathrm{ret}}(m)$, clean-release rate $P_{\mathrm{clean}}(m)$ and released-finger count; and (3) an instruction specifying the desired trade-off between grasp stability and downstream dexterity.

The instruction template provided to the LLM is:

\textit{
``You are selecting a finger-retention mask for a dual-task robotic manipulation problem. The robot hand has five fingers and must continue stabilizing the object from Task~A while freeing sufficient fingers to execute Task~B. Each candidate mask specifies which fingers remain in contact with the object and which fingers are released for downstream interaction.\\
A good mask should satisfy three objectives simultaneously: (1) maintain stable object retention, (2) allow clean finger release without disturbing the grasp, and (3) preserve dexterous and anatomically appropriate fingers for Task~B.\\
When evaluating candidates, consider not only the numerical metrics but also the semantic suitability of the finger assignment. For example, thumb-index opposition is generally preferable for stabilizing slender cylindrical objects, while releasing the thumb may significantly reduce grasp stability even if the clean-release score is high. Avoid overly conservative masks that release too few fingers and overly aggressive masks that sacrifice object stability.\\
Select the single mask that provides the best overall trade-off between grasp stability and downstream manipulation capability.''}

The LLM outputs the selected mask index, which is directly parsed and used as the execution mask $m^*$ during masked policy rollout. No reinforcement learning or fine-tuning is used for this allocation stage. Since all candidate masks and diagnostics are precomputed, the selection process is fully reproducible under deterministic decoding.

\noindent\textbf{Heuristic baseline.}
To evaluate whether language-based reasoning is necessary, we additionally compare against a heuristic baseline that selects the mask with the highest object-retention rate subject to a minimum-release constraint:
\begin{equation}
m_{\mathrm{heur}} = \arg\max_{m\in\mathcal{M}} P_{\mathrm{ret}}(m)
\quad \text{s.t.} \quad
5 - \|m\|_0 \ge 2.
\end{equation}

The constraint ensures that at least two fingers are released for Task~B execution. Unlike the LLM selector, this heuristic relies purely on release-test statistics and does not incorporate reasoning about grasp anatomy, object geometry, or downstream interaction requirements.

\noindent \textbf{Failure Case Analysis.}
Figure~\ref{fig:failure_llm} illustrates a representative example explaining why the LLM-based selector outperforms the heuristic baseline.
The heuristic chooses the thumb--index grasp because it achieves the highest object-retention rate among feasible masks.
While this decision is optimal for the current \textit{GraspBall} stage, it occupies the index finger, which is later required for manipulating the door.
As a result, the robot cannot establish a suitable contact configuration during \textit{OpenDoor}, leading to failure despite successfully retaining the object.
In contrast, the LLM tends to preserve fingers with higher functional utility, such as the index finger, which is commonly involved in precise contact-rich interactions.
Although this grasp provides slightly lower retention performance, it enables successful completion of both subtasks.
This example suggests that preserving functionally important fingers, rather than maximizing object retention alone, can lead to higher downstream task success.

\bibliography{example}  % .bib

\end{document}